\documentclass[11pt,a4paper]{article}
\usepackage[table,xcdraw]{xcolor}
\PassOptionsToPackage{numbers}{natbib}
\usepackage[hyperref]{acl2021}
\usepackage{times}
\usepackage{latexsym}

\usepackage{graphicx}

\usepackage{amsmath}
\usepackage{makecell}
\usepackage{tabularx}
\usepackage{booktabs}

\usepackage{arydshln}
\makeatletter
\g@addto@macro\normalsize{%
  \setlength\abovedisplayskip{2pt}
  \setlength\belowdisplayskip{2pt}
  \setlength\abovedisplayshortskip{2pt}
  \setlength\belowdisplayshortskip{2pt}
}

\makeatother

\definecolor{mygreen}{RGB}{0, 153, 0}
\definecolor{myorange}{RGB}{255, 153, 51}
\definecolor{mylightgreen}{RGB}{0, 204, 153}

\usepackage[font=small]{caption}

\usepackage{algorithm} 
\usepackage{algpseudocode} 

\usepackage{microtype}

\aclfinalcopy 

\title{Composable NLP Workflows for BERT-based Ranking and QA System}

\author{Gaurav Kumar* \\
  UC San Diego \\
  \texttt{gkumar@ucsd.edu} \\\And
  Murali Mohana Krishna Dandu* \\
  UC San Diego \\
  \texttt{mdandu@ucsd.edu}  \\}

\date{}

\begin{document}
\maketitle
\footnotetext{\label{code}Our code is open-source and is available at \href{https://github.com/gaurav5590/Doc-Ranker}{https://github.com/gaurav5590/Doc-Ranker} \\ * Equal contribution}
\begin{abstract}
There has been a lot of progress towards building NLP models that scale to multiple tasks. However, real-world systems contain multiple components and it is tedious to handle cross-task interaction with varying levels of text granularity. In this work, we built an end-to-end Ranking and Question-Answering (QA) system using Forte, a toolkit that makes composable NLP pipelines. We utilized state-of-the-art deep learning models such as BERT, RoBERTa in our pipeline, evaluated the performance on MS-MARCO and Covid-19 datasets using metrics such as BLUE, MRR, F1 and compared the results of ranking and QA systems with their corresponding benchmark results.
The modular nature of our pipeline and low latency of reranker makes it easy to build complex NLP applications easily.
\end{abstract}

\section{Introduction}
Building Natural Language Processing (NLP) applications involve multiple data systems and require several NLP components to be stitched together. For example, a QA pipeline will have the following components - query understanding, full-search from the entire document corpus, re-ranking to fine-tune a subset of results, identifying the answer phrase/sentence from the top documents, and finally showing the response in a consumable way to the end user. Such a complete end-to-end system requires multiple NLP techniques and may have different input output formats. To achieve this, we need a composable pipeline with standardized data formats coupled with the power of state-of-the-art NLP models.

The objective is to build and test an end-to-end Ranking and QA system utilizing Forte \cite{forte}, a toolkit for building composable NLP pipelines. We also wanted to build and test a real-world applicable Covid-19 QA system using our created pipeline which can be reused for further research and applications.

We have contributed during the project in the following ways:
\begin{itemize} 
  \item We built a Ranking and QA system using MS-MARCO Passage Ranking and QA datasets \cite{bajaj2018ms} and thoroughly evaluated each sub-component of it
  \item We built a Covid QA system which extends the previous pipeline and can be used in the real-world with few modifications
  \item We are contributing to open source technology Forte by adding the required missing processors and evaluators in the pipeline
  \item We provided examples for end-users to build a Retrieval and QA system with high quality code and documentation
\end{itemize}

\section{Forte}
Forte was build with one goal in mind - to standardize NLP interfaces. It is part of CASL (Composable Automatic and Scalable Learning) \cite{casl} open-source toolkit and provides functionalities to put together various NLP components together. Here are the major building blocks of Forte:
\begin{itemize} 
  \item \textbf{Data Structures \& Ontology:} Usually NLP applications need different information at different data granularity about the text pieces like named-entity mentions, POS tags, dependency links and some user defined groups. To support this, Forte has three template data types: Span (being, end), Link (parent, child) and Group (members) which are common across most of the NLP tasks. This supports interoperations across different NLP functions in the pipeline.  
  \item \textbf{DataPack:} DataPack contains all the information stored in the above datatypes and gets updated with new information as it passes through the pipeline. The elements of DataPack can be accessed at any stage of the pipeline using required granuarity and structure which perfectly supports cross-talk and interoperations 
  \item \textbf{Pipeline:} A pipeline is a collection of several re-usable readers, indexers, processors and evaluators. Each component in the pipeline will process the incoming DataPack and updates it's internal ontology structure
\end{itemize}

A clear understanding of the pipeline and the information flow will be clear through Figure 3 after which will be discussed in the next section.

\section{Ranking and QA System}
With the abundance of web documents, retrieving relevant and specific information from huge corpuses is crucial for lot of applications. Chatbots, Voice agents, Search engines - all of them use key components of ranking and QA. In light of the latest pandemic and the lack of unified information about Covid, we also wanted to build a retrieval system for Covid QA using CORD-19 dataset \cite{wang2020cord19}. However, since the dataset isn't suitable for evaluating intermediate systems, we first built and test the pipeline on MS-MARCO passage ranking and QA datasets.

\subsection{Components of the System}
\begin{figure}[h]
\includegraphics[width=0.45\textwidth]{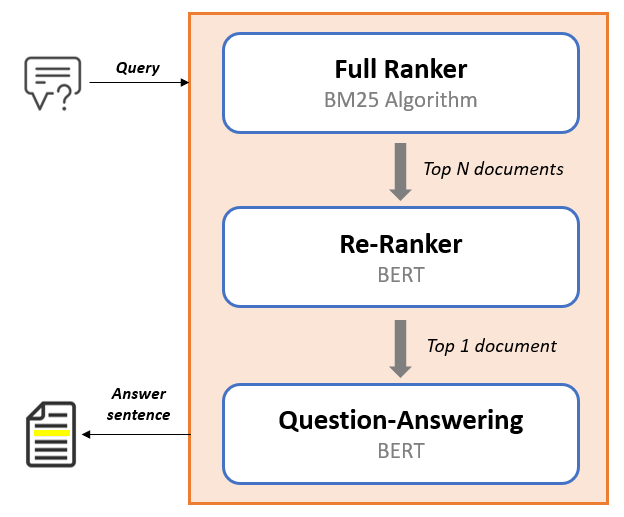}
\caption{\label{ranking-qa} Ranking and QA System}
\end{figure}

An end-to-end QA system consists of three major components as shown in Figure \ref{ranking-qa}.
\begin{itemize} 
  \item \textbf{Full-Ranker:} A system which ranks the entire corpus using a simple and faster algorithm based on keywords. The goal  is to ideally have 100\% retrieval recall and fetches top-N documents with very less latency. We have used Okapi-BM25 algorithm \cite{bm25} for the full-ranking and have tested the system with multiple top-N documents
  \item \textbf{Re-Ranker:} The top-N documents fetched from the Full-ranker are ranked again using an advanced model like BERT \cite{devlin2019bert}. Here, the ranking will be fine-tuned using the query-document feature representations learnt by the model. We have used a HuggingFace BERT model \cite{nogueira2020passage} trained on MS-MARCO itself where it will provide a score for a query-document combination
  \item \textbf{Question-Answering:} The top-1 document scored by the re-ranker will be used as a context text for MRC-style QA \cite{rajpurkar2016squad}. A specific phrase in the document is selected by the model as the answer span and we took the entire sentence of the document for evaluation and easier consumption. We have used a QA BERT model \cite{hugging_qa} trained on SQUAD data.
  \end{itemize}
  
\subsection{Complete Pipeline}
\begin{figure}[h]
\centering
\includegraphics[width=0.5\textwidth]{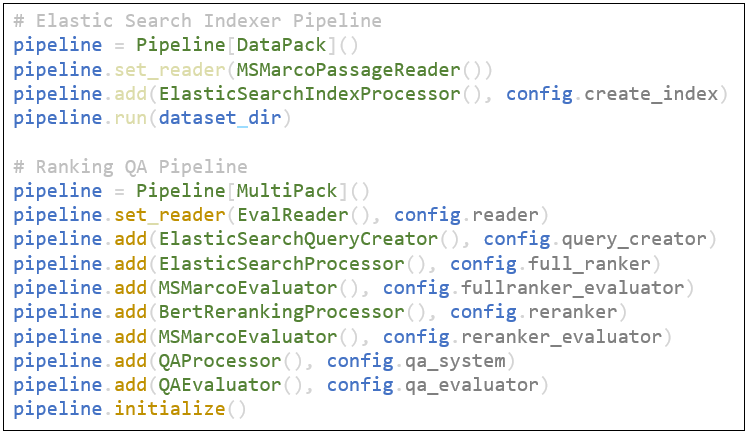}
\caption{\label{forte}Forte Pipeline}
\end{figure}

The code pipeline in Figure \ref{forte} consists of a series of Forte processors where the data flows from one to next and the internal MultiPack gets modified accordingly. Here is the code snippet for indexing, ranking, answering and evaluating the results for a given set of user queries.

\begin{figure*}[h]
\centering
\includegraphics[width=0.98\textwidth]{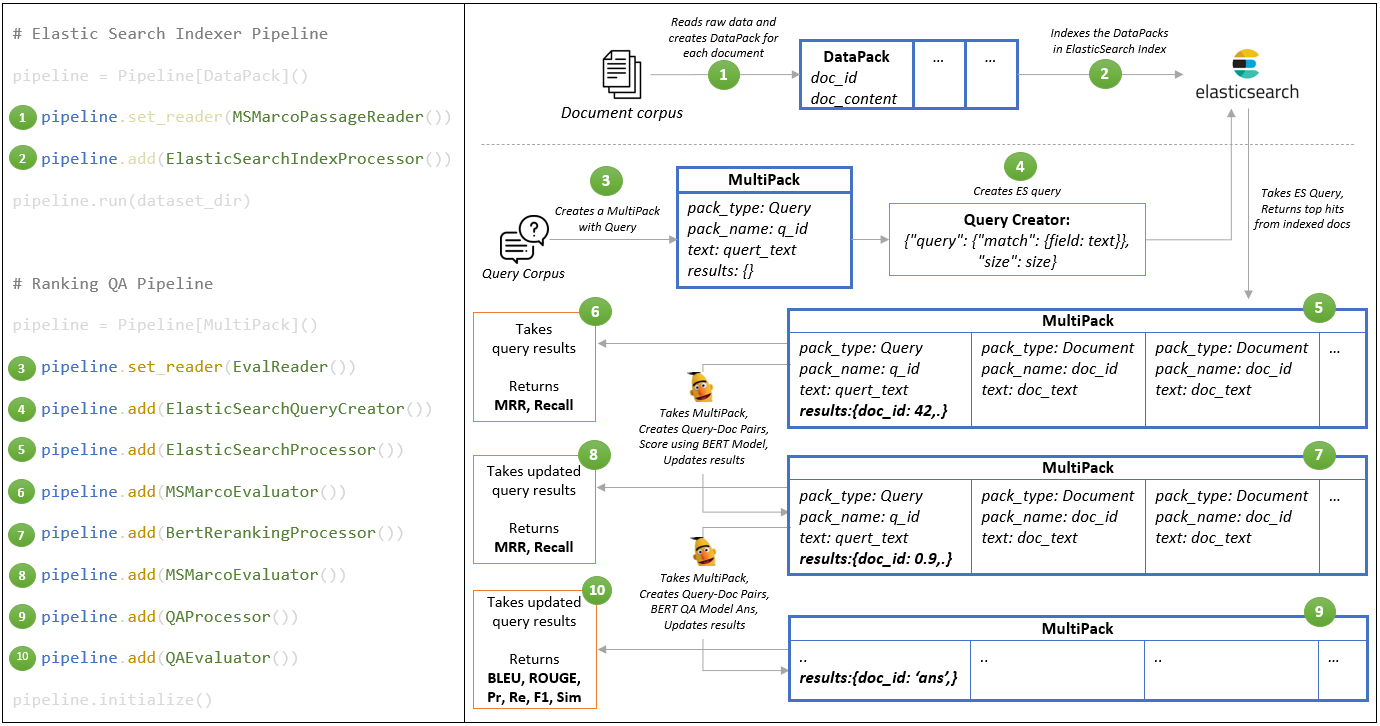}
\caption{\label{info-flow}Information flow across the Forte pipeline}
\end{figure*}

Figure \ref{info-flow} provides the complete picture on how the data flows through each component of the pipeline combining readers, ranking and QA processors and evaluators. You can also observe that the use of DataPack and MultiPack makes it easier for information flow and access at each stage.

As you can see, such a pipeline makes it easier to build similar QA systems. For example, by changing the reader to CordReader() with the relevant datasets provided in the config.yml, we have created a Covid QA. We will mention about the ohter engineering changes will discussing the results.

\section{Results and Discussion}

\subsection{Overview}
We have used MS-MARCO passage ranking and QA dataset for building the complete system. The dataset contains 8.1 million passages collected from real Bing questions and documents. We have used 1000 to 7000 queries from the development set to test our pipeline and showcase the results. Here are the different evaluation metrics used to test the systems:
\begin{itemize}
    \item \textbf{MRR@N:} Mean-Reciprocal-Rank is the average of multiplicative inverse of the rank of the first correct answer. It measures where exactly the relevant document resides in the top-N ranking
    \item \textbf{Recall@N:} Recall ignores the position and checks whether the relevant document is present in the top-N. This is useful when you have multiple ranking systems and wants to maintain high-recall in the full-ranking search results
    \item \textbf{BLEU-1:} BLEU score measures the precision between reference and predicted text at uni-gram level
    \item \textbf{ROUGE-L:} ROUGE is a recall oriented metric that measures the longest common subsequence between two pieces of text
    \item \textbf{F1:} F1 is calculated using the total uni-gram tokens that are matched
    \item \textbf{Semantic Similarity:} Measures SpaCy's similarity which is computed through cosine of word representations (using GloVe etc.)
\end{itemize}

\begin{table}[h]
\centering
\begin{tabular}{llc}
\hline  
\textbf{System} & \textbf{Metric}     & \textbf{Value} \\ \hline
Full-Ranking    & MRR@10              & 0.16           \\
                & Recall@10           & 34\%           \\ \hline
Re-Ranking      & MRR@10              & 0.34           \\
                & Recall@10           & 58\%           \\ \hline
QA              & BLEU-1              & 0.32           \\
                & ROUGE-L             & 0.32           \\
                & F1 (Tokens Match)   & 0.32           \\
                & Semantic Similarity & 80\%           \\ \hline
\end{tabular}
\caption{\label{overall-table} Overall Metrics}
\end{table}

\begin{table*}[h]
\resizebox{\textwidth}{!}{%
\begin{tabular}{|r|r|r|r|r|r|r|r|r|r|}
\hline
\multicolumn{1}{|l|}{}             & \multicolumn{1}{l|}{}                     & \multicolumn{4}{c|}{\cellcolor[HTML]{F9CB9C}\textbf{Full-Ranker}}                                                                                                                                                                                                & \multicolumn{4}{c|}{\cellcolor[HTML]{CFE2F3}\textbf{Re-ranker}}                                                                                                                                                                                                    \\ \hline
\multicolumn{1}{|l|}{\textbf{Re-Ranking Size}} & \multicolumn{1}{l|}{\textbf{Time per query (s)}} & \multicolumn{1}{l|}{\cellcolor[HTML]{F9CB9C}\textbf{MRR@10}} & \multicolumn{1}{l|}{\cellcolor[HTML]{F9CB9C}\textbf{MRR@100}} & \multicolumn{1}{l|}{\cellcolor[HTML]{F9CB9C}\textbf{Recall@10}} & \multicolumn{1}{l|}{\cellcolor[HTML]{F9CB9C}\textbf{Recall@100}} & \multicolumn{1}{l|}{\cellcolor[HTML]{CFE2F3}\textbf{MRR@10}} & \multicolumn{1}{l|}{\cellcolor[HTML]{CFE2F3}\textbf{MRR@100}} & \multicolumn{1}{l|}{\cellcolor[HTML]{CFE2F3}\textbf{Recall@10}} & \multicolumn{1}{l|}{\cellcolor[HTML]{CFE2F3}\textbf{Recall@100}} \\ \hline
1                                                & 0.48                                       & 0.09                                                         & 0.09                                                          & 0.09                                                            & 0.09                                                             & 0.09                                                         & 0.09                                                          & 0.09                                                            & 0.09                                                             \\ \hline
10                                               & 0.54                                      & 0.16                                                         & 0.16                                                          & 0.34                                                            & 0.34                                                             & 0.23                                                         & 0.23                                                          & 0.34                                                            & 0.34                                                             \\ \hline
50                                               & 0.85                                     & 0.16                                                         & 0.17                                                          & 0.34                                                            & 0.50                                                             & 0.28                                                         & 0.28                                                          & 0.45                                                            & 0.50                                                             \\ \hline
100                                              & 1.24                                      & 0.16                                                         & 0.17                                                          & 0.34                                                            & 0.59                                                             & 0.30                                                         & 0.30                                                          & 0.50                                                            & 0.59                                                             \\ \hline
500                                              & 4.61                                      & 0.16                                                         & 0.17                                                          & 0.34                                                            & 0.59                                                             & 0.33                                                         & 0.33                                                          & 0.56                                                            & 0.73                                                             \\ \hline
\textbf{1000}                                    & \textbf{8.81}                            & \textbf{0.16}                                                & \textbf{0.17}                                                 & \textbf{0.34}                                                   & \textbf{0.59}                                                    & \textbf{0.34}                                                & \textbf{0.35}                                                 & \textbf{0.58}                                                   & \textbf{0.77}                                                    \\ \hline
\end{tabular}%
}
\caption{\label{ranker-res-table}Full-Ranking and Re-Ranking results for 1000 queries}
\end{table*}

\begin{table*}[h]
\resizebox{\textwidth}{!}{%
\begin{tabular}{|r|r|r|r|r|r|r|r|r|r|}
\hline
\multicolumn{1}{|l|}{}             & \multicolumn{9}{c|}{\cellcolor[HTML]{B6D7A8}\textbf{QA}}                                                                                                                                                                                                                                                                                                                                                                                                                                                                                                                                   \\ \hline
\multicolumn{1}{|l|}{\textbf{Re-Ranking Size}} & \multicolumn{1}{l|}{\cellcolor[HTML]{B6D7A8}\textbf{BLEU-1}} & \multicolumn{1}{l|}{\cellcolor[HTML]{B6D7A8}\textbf{BLEU-2}} & \multicolumn{1}{l|}{\cellcolor[HTML]{B6D7A8}\textbf{BLEU-3}} & \multicolumn{1}{l|}{\cellcolor[HTML]{B6D7A8}\textbf{BLEU-4}} & \multicolumn{1}{l|}{\cellcolor[HTML]{B6D7A8}\textbf{ROUGE-L}} & \multicolumn{1}{l|}{\cellcolor[HTML]{B6D7A8}\textbf{PRECISION}} & \multicolumn{1}{l|}{\cellcolor[HTML]{B6D7A8}\textbf{RECALL}} & \multicolumn{1}{l|}{\cellcolor[HTML]{B6D7A8}\textbf{F1}} & \multicolumn{1}{l|}{\cellcolor[HTML]{B6D7A8}\textbf{Semantic Sim}} \\ \hline
1                                                & 0.24                                                         & 0.15                                                         & 0.11                                                         & 0.09                                                         & 0.22                                                          & 0.20                                                            & 0.23                                                         & 0.21                                                     & 0.75                                                               \\ \hline
10                                               & 0.30                                                         & 0.21                                                         & 0.17                                                         & 0.15                                                         & 0.29                                                          & 0.26                                                            & 0.32                                                         & 0.29                                                     & 0.79                                                               \\ \hline
50                                               & 0.31                                                         & 0.23                                                         & 0.19                                                         & 0.17                                                         & 0.31                                                          & 0.27                                                            & 0.34                                                         & 0.30                                                     & 0.79                                                               \\ \hline
100                                              & 0.31                                                         & 0.24                                                         & 0.20                                                         & 0.18                                                         & 0.32                                                          & 0.28                                                            & 0.35                                                         & 0.31                                                     & 0.80                                                               \\ \hline
500                                              & 0.32                                                         & 0.24                                                         & 0.21                                                         & 0.19                                                         & 0.32                                                          & 0.29                                                            & 0.36                                                         & 0.32                                                     & 0.80                                                               \\ \hline
\textbf{1000}                                    & \textbf{0.32}                                                & \textbf{0.25}                                                & \textbf{0.21}                                                & \textbf{0.19}                                                & \textbf{0.32}                                                 & \textbf{0.29}                                                   & \textbf{0.36}                                                & \textbf{0.32}                                            & \textbf{0.80}                                                      \\ \hline
\end{tabular}%
}
\caption{\label{qa-res-table}QA results on 1000 queries}
\end{table*}

Table \ref{overall-table} shows the above metrics for different systems. We have achieved a Full-Ranking MRR@10 of 0.16 and Recall@10 of 34\%. Note that we have used 1000 full-ranking size for the overall metrics. As we go to the re-ranker, we can observed that MRR@10 doubles and Recall@10 almost improves by 80\%. These re-ranking metrics are in a very similar range of MS-MARCO re-ranking leaderboard metrics. \cite{msmarco_leader}. Going to the QA, we see that BLEU, ROGUE and F1 are in the range of 0.32. The MS-MARCO QA leaderboard scores \cite{msmarco_leader} are in range of 0.50 and the difference is due to the error propagation from the upstream rankers. For example, we see that the re-ranking recall is 58\% and hence the QA metrics will be atleast down by 40\% compared to the standalone QA benchmarking tasks. Also, note that we are using the top-1 re-ranked document for QA and increasing the search size for QA will definitely improve the results. MRR of 0.34 implies that we the correct document is on average in top-3 and search for answer in top-3 documents will definitely improve the QA results.

\subsection{Sample Results}
Table \ref{sample-table} show some sample results of the pipeline with re-ranking size of 1000. We can see that the system is able to handle different types of question structures such as 'what', 'who' as well as phrasal questions like 'tristesse definition' which are commonly used in search engines.

\begin{table}[hbt!]
    \small
    \centering
    \begin{tabular}{m{0.45\textwidth}}
        \hline\hline
         \textbf{Q: } What is priority pass \\
         \textbf{A: }Priority Pass is an independent airport lounge access program. \\ \hline
         
         \textbf{Q: } tristesse definition \\
         \textbf{A: } Tristesse is a French word meaning sadness. \\ \hline
         
          \textbf{Q: } tricuspid atresia definition \\
         \textbf{A: }   Tricuspid atresia is a type of heart disease that is present at birth (congenital heart disease), in which the tricuspid heart valve is missing or abnormally developed. \\ \hline
         
          \textbf{Q: } who proposed the geocentric theory \\
         \textbf{A: }  The geocentric model, also known as the Ptolemaic system, is a theory that was developed by philosophers in Ancient Greece and was named after the philosopher Claudius Ptolemy who lived circa 90 to 168 A.D. \\ \hline \hline
         
    \end{tabular}
    \caption{\label{sample-table}QA results for sample queries}
    \label{samples}
\end{table}

\subsection{Re-Ranking Size}
Re-ranker is a much costlier operation since it uses bigger BERT architectures. You can see in Table \ref{ranker-res-table} that the retrieval per query increases with the size of documents we are re-ranking. Although we optimized the BERT re-ranking using GPU-batching, we would like to see how the results vary with different search sizes. If you observe the re-ranking results across different re-ranking sizes, there is a consistent improvement till 100-500 size and the returns get diminished after that. At size 500, we achieve very close results of 0.33 and 56\% of MRR@10 and Recall@10 compared to 0.34 and 56\% for size 1000 with a reduction in retrieval time by ~50\%. 

When we look at the similar Table \ref{qa-res-table} for QA results, we observe that all the results in terms of BLEU, ROUGE and F1 get saturated after 50-100 size. Combining the results from these two tables, we can say that getting top ~100 size for re-ranking would be ideal for this experimental setup.

\begin{table*}[t]
\resizebox{\textwidth}{!}{%
\begin{tabular}{|l|l|r|r|r|r|r|r|r|r|r|}
\hline
&                             & \multicolumn{9}{c|}{\cellcolor[HTML]{B6D7A8}\textbf{QA}}                                                                               \\ \hline
\textbf{Re-Ranking Size} & \textbf{Time per query (s)} & \multicolumn{1}{l|}{\cellcolor[HTML]{B6D7A8}\textbf{BLEU-1}} & \multicolumn{1}{l|}{\cellcolor[HTML]{B6D7A8}\textbf{BLEU-2}} & \multicolumn{1}{l|}{\cellcolor[HTML]{B6D7A8}\textbf{BLEU-3}} & \multicolumn{1}{l|}{\cellcolor[HTML]{B6D7A8}\textbf{BLEU-4}} & \multicolumn{1}{l|}{\cellcolor[HTML]{B6D7A8}\textbf{ROUGE-L}} & \multicolumn{1}{l|}{\cellcolor[HTML]{B6D7A8}\textbf{PRECISION}} & \multicolumn{1}{l|}{\cellcolor[HTML]{B6D7A8}\textbf{RECALL}} & \multicolumn{1}{l|}{\cellcolor[HTML]{B6D7A8}\textbf{F1}} & \multicolumn{1}{l|}{\cellcolor[HTML]{B6D7A8}\textbf{Semantic Sim}} \\ \hline
\multicolumn{1}{|r|}{100}  & \multicolumn{1}{r|}{1.21}   & \cellcolor[HTML]{F8F8F8}{\color[HTML]{1D1C1D} 0.20}          & \cellcolor[HTML]{F8F8F8}{\color[HTML]{1D1C1D} 0.15}          & 0.13                                                         & 0.12                                                         & 0.22                                                          & 0.18                                                            & 0.29                                                         & 0.22                                                     & 0.71                                                               \\ \hline
\multicolumn{1}{|r|}{1000} & \multicolumn{1}{r|}{6.64}   & 0.20                                                         & 0.15                                                         & 0.13                                                         & 0.12                                                         & 0.22                                                          & 0.18                                                            & 0.29                                                         & 0.22                                                     & 0.71                                                               \\ \hline
\end{tabular}%
}
\caption{\label{covid-results}QA results for Cord-19 dataset on 2019 queries}
\end{table*}

\subsection{Covid QA}
Given the wide-spread pandemic and the public curiosity to understand it better, we wanted to build a COVID QA system by tweaking the above pipeline. We have used the CORD-19 dataset \cite{wang2020cord19} as our document corpus, and the COVID-QA dataset \cite{moller-etal-2020-covid} to test the results. Note that the Covid QA dataset has been created using only 147 documents out of 200K+ CORD-19 documents.

Since the CORD-19 documents are very huge and the pre-trained BERT models under use are restricted by 512 token input size, we pre-processed the documents before indexing the documents. We followed the chunking and striding approach \cite{zhang2020covidex} to chunk our documents into smaller texts, which is a common strategy while using such BERT models. We chunked the long text by every 60 tokens keeping an overlapping stride of 15 tokens. With this, the total 212K documents have become around 18M short text passages as an input to our full-search. We have used the deepset.ai Covid model \cite{roberta} for QA but used the same previous MS-MARCO model for re-ranking.

Additionally, Covid-QA is SQUAD style dataset where the reference answers were short phrases (given the context) and is usually evaluated by F1 and exact match. However, we have used the full-sentence as answers and used other evaluation metrics as well along with F1.

Table \ref{covid-results} shows the results of the QA system where the BLEU-1, ROUGE-L and F1 are in the range of 0.22 with Semantic Similarity being in range of 0.71. We can also see that there is no difference using higher re-ranking size (100 vs 1000), implying that the re-ranker model is not adding much value to this particular task. It is due to the fact that we are using a generic BERT ranker model which will not have technical vocabulary common in CORD dataset. Hence training and fine-tuning a model to this dataset would be an ideal next step to improve the results. \cite{ngai2021transformerbased} shows that they the benchmark F1-score is in the range of 0.25-0.30 showing that our pipeline is not far away to achieve better results.

Table \ref{covid-sample} shows some sample queries along with both ground truth and predicted answers from our pipeline. As you can see, it is able to identify the sentence containing the answer.

\begin{table}[hbt!]
    \small
    \centering
    \begin{tabular}{m{0.45\textwidth}}
        \hline\hline
         \textbf{Q: } What is the main cause of HIV-1 infection in children? \\
         \textbf{G: } Mother-to-child transmission (MTCT) is the main cause of HIV-1 infection in children worldwide. \\
         \textbf{A: } Mother-to-child transmission (MTCT) is the main cause of HIV-1 infection in children worldwide. \\ \hline
         
         \textbf{Q: } What is the size of bovine coronavirus? \\
         \textbf{G: } 31 kb \\
         \textbf{A: } The BCoV is a RNA virus, nonenveloped, diameter of 120 nm, single-stranded (ssRNA) positive-sense, and non-segmented with 27-32 Kb size (ICTV 2015) \\ \hline
         
          \textbf{Q: } How long is the SAIBK gene? \\
        \textbf{G: } 27,534 nucleotides \\
         \textbf{A: } The complete genome of the SAIBK strain is 27,534 nucleotides (nt) in length, including the poly(A) tail. \\ \hline 
         
        \textbf{Q: } What does the hamster model for HCPS caused by? \\
        \textbf{G: } by capillary leak that results in pulmonary edema and the production of a pleural effusion with exudative characteristics\\
         \textbf{A: } the hamster model for HCPS appears to be caused by capillary leak that results in pulmonary edema and the production of a pleural effusion with exudative characteristics.\\ \hline \hline 
         
    \end{tabular}
    \caption{\label{covid-sample}QA results for sample queries of CORD dataset. Q: question, G: ground truth answer, A: predicted answer}
    \label{samples}
\end{table}

\section{Conclusion and Future Work}
In this project, we build a composable ranking and QA pipeline with comparable results on MS MARCO and Covid-19 QA datasets. The modular nature of our pipeline as depicted in Figure \ref{forte} makes it easy to build complex NLP applications easily. In future, we would like to train and fine-tune the deep learning ranking models behind our processors for specific datasets. Also, decreasing the latency of re-ranker while maintaining the accuracy will further help in real-world search applications. Covid-QA ranking performance can be further evaluated by utilizing reference context on top of the existing metrics.

\section{Acknowledgements}
We would like to thank Dr. Zhiting Hu and Dr. Zhengzhong (Hector) Liu for their valuable inputs towards this work. We also extend our gratitude towards Petuum Inc. for providing us the computing support needed to run our pipeline on GPU.

\bibliographystyle{unsrt}
\bibliography{acl2021}


\end{document}